\documentclass[conference]{IEEEtran}

\usepackage{amsmath,amssymb,amsfonts}
\usepackage{algorithmic}
\usepackage{graphicx}
\usepackage{textcomp}
\usepackage{xcolor}
\usepackage{amsmath}
\usepackage{float}
\usepackage{algorithm}
\usepackage{algorithmic}
\usepackage{booktabs}
\usepackage{multirow}
\usepackage{nomencl}
\usepackage{tikz}
\usepackage{pgfplots}
\usetikzlibrary{shapes, arrows}
\usepackage{todonotes}
\usepackage{svg}
\usepackage{subfig}
\renewcommand{\baselinestretch}{0.925}
\newbox{\bigpicturebox}

\begin{document}
\title{Exploring Lightweight Federated Learning for Distributed Load Forecasting
}

\author{\IEEEauthorblockN{Abhishek Duttagupta, Jin Zhao, Shanker Shreejith}
\IEEEauthorblockA{ Electronic \& Electrical Engineering\\
Trinity College Dublin, Ireland\\
Email: \{aduttagu,zhaoj6,shankers\}@tcd.ie}}

\maketitle

\begin{abstract}
Federated Learning (FL) is a distributed learning scheme that enables deep learning to be applied to sensitive data streams and applications in a privacy-preserving manner. This paper focuses on the use of FL for analyzing smart energy meter data with the aim to achieve comparable accuracy to state-of-the-art methods for load forecasting while ensuring the privacy of individual meter data.
We show that with a lightweight fully connected deep neural network, we are able to achieve forecasting accuracy comparable to existing schemes, both at each meter source and at the aggregator, by utilising the FL framework.
The use of lightweight models further reduces the energy and resource consumption caused by complex deep-learning models, making this approach ideally suited for deployment across resource-constrained smart meter systems. 
With our proposed lightweight model, we are able to achieve an overall average load forecasting RMSE of 0.17, with the model having a negligible energy overhead of 50 mWh when performing training and inference on an Arduino Uno platform.

\end{abstract}


\section{Introduction} 
Electricity demand forecasting is a crucial function in the energy industry to allow both producers and distribution system operators to optimize the generation and distribution of electricity in real-time~\cite{weron2014electricity}. 
It aims to balance the supply and demand curve and maintain the stability of the electric grid across the entire network of producers and consumers. 
Smart energy meters~\cite{arif2013experimental} play a vital role in load forecasting by integrating communication capabilities on top of their ability to monitor energy consumption accurately~\cite{spano2014last}. 
Smart meters transmit demand data from consumers at specified intervals to distribution system operators, allowing them to observe the data in real time, make predictions on future demand, and adapt the grid to respond to this expected demand. 
This data-driven approach has enabled electric grid operators to optimize energy generation and storage to adapt to load peaks from the supply-side, and at the same time achieve demand-side load balancing through novel schemes like dynamic pricing tariff rates to achieve supply-demand equilibrium~\cite{diaz2019prediction}. 

Researchers have proposed different algorithms for load forecasting that primarily enables dynamic energy pricing and usage optimization \cite{khan2016load}. 
In~\cite{diaz2019prediction}, the authors apply gradient-boosted regression trees on the Spanish Energy dataset to minimize prediction error for a day-ahead forecast, showing the effectiveness of ML models in load forecasting. 
However, load forecasting is a challenging task due to the stochastic nature of customer usage profiles. 
In~\cite{tascikaraoglu2016short}, the authors show that the inclusion of spatial and temporal parameters can have a significant impact on the accuracy of load prediction. 
Area-based clustering techniques have also been explored to group customers with similar electric energy consumption patterns in a spatiotemporal window to improve the accuracy of load forecasting~\cite{bandara2020forecasting,quilumba2014using}.

Hybrid models combining classical and deep learning methods have also demonstrated promising results in load forecasting. 
In~\cite{saxena2019hybrid}, the authors developed a hybrid model utilizing ARIMA, logistic regression, and deep neural networks to predict peak load days with a 70~\% accuracy. 
Other studies reported in the literature have proposed models based on Convolutional neural networks (CNNs) with fuzzy time series, achieving superior results compared to traditional models~\cite{sideratos2020novel,sadaei2019short}. 
Using complex models such as a standalone Long Short Term Memory (LSTM) model~\cite{memarzadeh2021short}, a combination of LSTM and CNN~\cite{guo2020short}, and Gated Re-current Units (GRU) combined with CNN~\cite{sajjad2020novel} have also shown to achieve similar prediction performance. 
The use of rolling updates and Bi-LSTMs was shown to reduce the computational time over other deep learning methods, while also improving the accuracy of load forecasting~\cite{wang2019bi}. 

While most deep learning models have been primarily focused on improving forecasting performance, privacy concerns have risen with the centralized learning approach used by load forecasting schemes. 
In the case of centralized models, data from smart meters of consumers is monitored and accumulated at locations within the distribution grid, which is used by the model to predict peak loads or other learned parameters about the demand in specific clusters. 
It has been shown that non-intrusive analysis of granular load curves can provide information about appliances, usage patterns and hence personal details about consumers~\cite{6990609,5874919}.
Privacy-preserving machine learning (ML) through federated learning (FL) is a promising scheme to address the challenges of centralized learning on sensitive data~\cite{5874919}. 
With FL, individual models can be trained collaboratively and locally, removing the need to accumulate and share sensitive data across the network. 
While FL has shown to be effective in other domains~\cite{hu2018federated, pfohl2019federated}, its application to load forecasting is very limited. 
In~\cite{GHOLIZADEH2022100470}, the authors propose a combination of LSTM and dense neural network within an FL framework for load forecasting and utilize clustering to improve the model's prediction performance. 
However, the root mean squared error (RMSE) measure of the federated approach was higher than the centralized learning technique. 
The work in~\cite{savi2021short} achieved better prediction performance using a global generalized model aggregated from locally trained LSTM models at the consumer end, and by incorporating spatiotemporal factors into the framework. 
However, the complexity of local and global models as well as their high energy consumption is a barrier to their widespread deployment in constrained devices such as smart meters.

In this work, we apply an FL framework for short-term load forecasting using the openly available smart meter energy consumption dataset for London households~\cite{londondataset}. 
The key objective is to show that comparable accuracy to state-of-the-art methods in load prediction can be achieved using lightweight deep neural network (DNN) models within an FL framework to restrict the need for granular data sharing and aggregation across the network. 
To replicate real-world non-independent and identical distribution (i.i.d.) conditions, we consider limited random connections for each round of FL, simulating scenarios where not all devices are participating in the learning process. 
The main contributions of this paper are summarized below.
\begin{enumerate}
    \item A lightweight feed-forward model within an FL framework to achieve identical and/or superior short-term load forecasting performance, even while considering limited device connections within the cluster (non i.i.d condition).
    \item Combining the FL flow with clustering to achieve a better global representation within each cluster, leading to improved forecasting performance. 
    \item Evaluate our approach and demonstrate that lightweight models are sufficient to achieve similar load forecasting performance when compared to other complex models.
    \item Quantify the overheads of the proposed lightweight models on a lightweight microcontroller that mimics an actual smart meter device.
\end{enumerate}

The rest of this paper is organized as follows: Section II discusses the proposed federated learning framework, and Section III illustrates the employed methodology in detail; Section IV presents the case study performed with the results analysis; and conclusions are drawn in Section V.

\section{Light-weighted FL-based Load Forecasting Scheme }

\subsection{Federated Learning Algorithm}

\begin{figure}[t!]
    \centering
        \centering
        \includegraphics[width=0.7\linewidth]{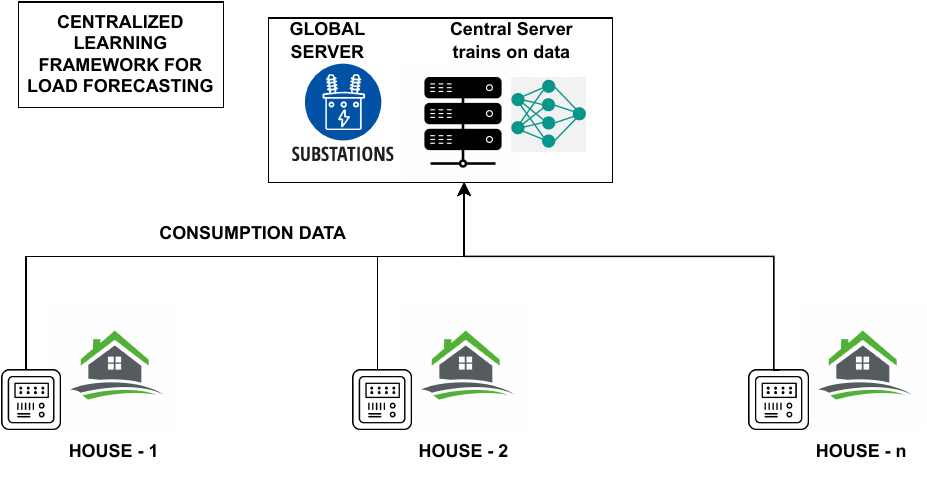}
        \caption{Traditional centralized setup for load forecasting.}
        \label{fig:traditional}
    \vspace{-2mm}
\end{figure}

\begin{figure}
        \centering
        \includegraphics[width=0.7\linewidth]{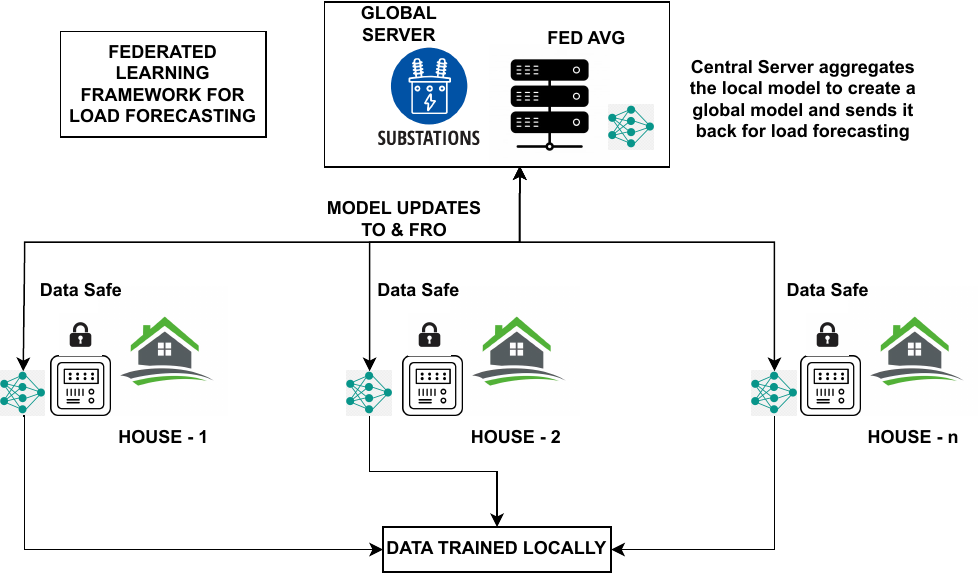}
        \caption{Federated learning setup for load forecasting.}
        \label{fig:federated}
    \vspace{-2mm}
\end{figure}

In a traditional centralized scenario \ref{fig:traditional}, energy consumption data from participating household smart meters is sent to a cluster head (typically at a substation level), where the global server performs model training and inference. 
In contrast, with an FL setup \ref{fig:federated}, each household performs its own edge training of local models based on their individual consumption data. These updated models are then sent to the global server at the substation for aggregation, creating the global model. The global model is subsequently shared with the households for evaluation.
The entire cycle keeps repeating, allowing the local and global model to evolve based on changes in individual consumers' consumption and generation. 


The federated algorithm flow \ref{alg:federated_learning} used for our process is derived, from the work in~\cite{mcmahan2017communication}. 
The algorithm begins by selecting a set of user devices (N=0,1,2,...,n) from the total device set (M) and initializing an initial global model (W), which is then passed to these devices. 
Training is performed in parallel on these selected devices using their local data (P), considering batch sizes (B), learning rate ($\eta$), and local epochs (E). 
After training, the updated models (w) from these devices are shared with the substation, where federated averaging occurs for server aggregation, resulting in a global model (W). 
The global model is then shared back with all the devices, which update their models accordingly, and the process is repeated until convergence is reached. 
By Convergence, we refer to the point where no noticeable improvement in the global loss function (GLF) occurs with increasing federated training rounds, indicating that the GLF has reached a minimum. 

\begin{algorithm}[t!]
\begin{algorithmic}[1]
\small
\REQUIRE Total device set $M$, selected device set $N$, batch size $B$, learning rate $\eta$, local epochs $E$
\ENSURE Global model $W$
\STATE Initialization: Randomly initialize global model $W$
\WHILE{not converged}
\FOR{each device $n$ in selected device set $N$}
\STATE Retrieve local data $P$ on device $n$
\STATE Train local model $w_n$ on data $P$ for $E$ epochs with batch size $B$ and learning rate $\eta$
\STATE Share local model $w_n$ with central server
\ENDFOR
\STATE Aggregate models $w_n$ from all devices on a central server using federated averaging
\STATE Update global model $W$ as the average of all local models $w_k$
\STATE Share global model $W$ with all devices
\ENDWHILE
\end{algorithmic}
\caption{Federated learning algorithm}
\label{alg:federated_learning}
\end{algorithm}

Adopting the representation in~\cite{shen2023share} for our case, an FL cluster with \emph{n} clients communicating with the global server(substation for our case), if each client has a \emph{k} fixed number of data points in its respective dataset, the input/output relation for a given local dataset \emph{S} for a client \emph{i} can be represented by the equation eq.~\ref{eq1}. Here \emph{x} and \emph{y} represent the input and output vectors, respectively.

\begin{equation}
\mathcal{S}_i = \{(x_{i}, y_{i})\}_{i=1}^{k_i} \label{eq1}
\end{equation}

The learning process can hence be captured using the model f(.; $\theta$) that maps an input \emph{x} to the predicted label \emph{y}. 
Here, $\theta$ represents the trainable parameter of the model f. 
Assuming a loss function of \emph{L} for this model capturing the error in the prediction  f(x; $\theta$) given the true label y, the local objective of the client \emph{i} is given by the equation eq.~\ref{eq2} and the global server objective in an FL setting can be represented by the equation eq.~\ref{eq3}.  

\begin{equation}
L(\theta; \mathcal{S}_i) = \frac{1}{k_i} \sum_{i=1}^{k_i} l(f(x_{i}; \theta), y_{i}) \label{eq2}
\end{equation}

\begin{equation}
\arg \min_{\theta} L(\theta) = \frac{1}{n} \sum_{i=1}^{n} L(\theta; \mathcal{S}_i) \label{eq3}
\end{equation}

\subsection{Performance Metrics}

With errors varying according to household average consumption level, MAE (calculated using eq.~\ref{mae eq}) and RMSE (given by eq.~\ref{rmse eq}) are not ideal metrics to accurately assess short-term load forecasting at small energy consumption units. Mean absolute percentage error (MAPE, given by eq.~\ref{mape eq}) provides a more reliable measure by capturing the absolute percentage deviation between predicted and actual values. Using MAPE with MAE and RMSE allows for a comprehensive evaluation and comparison of prediction accuracy independent of a household's average consumption.

\begin{equation}
MAE = \frac{1}{N} \sum_{i=1}^{N} |X(i) - Y(i)| \label{mae eq}
\end{equation}

\begin{equation}
RMSE = \sqrt{\frac{1}{N} \sum_{i=1}^{N} [X(i) - Y(i)]^2} \label{rmse eq}
\end{equation}

\begin{equation}
MAPE = \frac{1}{N} \sum_{i=1}^{N} \left| \frac{X(i) - Y(i)}{X(i)} \right| \times 100 \label{mape eq}
\end{equation}

As before, in the case of equations eq.~\ref{mae eq} to eq,~\ref{mape eq}, X represents the actual value and Y represents the predicted value, with N as the number of data points.

\section{DISTRIBUTED LOAD FORECASTING METHODOLOGY}

For our simulation and testing, we utilized the smart meter energy consumption dataset from London Households~\cite{londondataset}. 
This dataset comprises energy consumption data recorded from smart meters installed in 5,567 households across London. 
The data covers the period from November 2011 to February 2014, with half-hourly energy consumption data, containing four columns: a unique tag, tariff type (standard or dynamic pricing), time-stamp, and half-hourly electric consumption (in kWh).

\begin{figure}[t!]
    \centering
        \centering
        \includegraphics[width=0.85\linewidth]{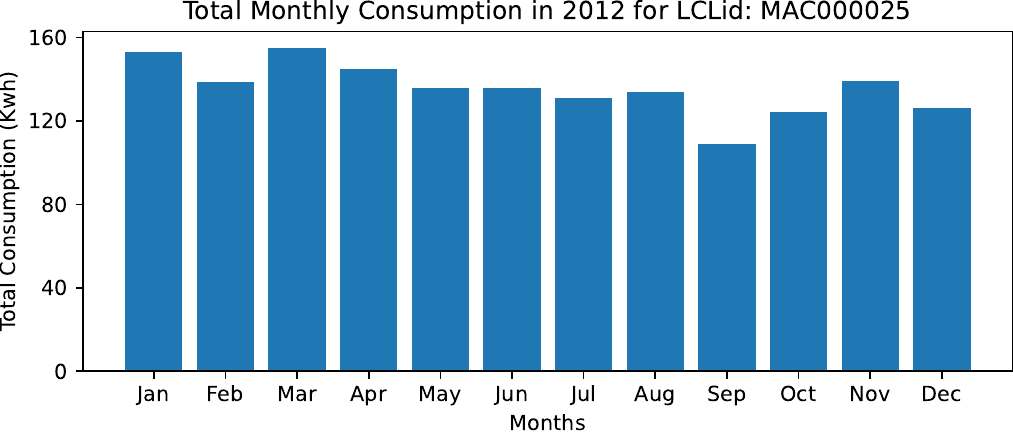}
        \caption{Energy profile yearly trend of a random household}
        \label{subfig:p3}
    \vspace{-2mm}
\end{figure}

\begin{figure}[t!]
    \centering
        \centering
        \includegraphics[width=\linewidth]{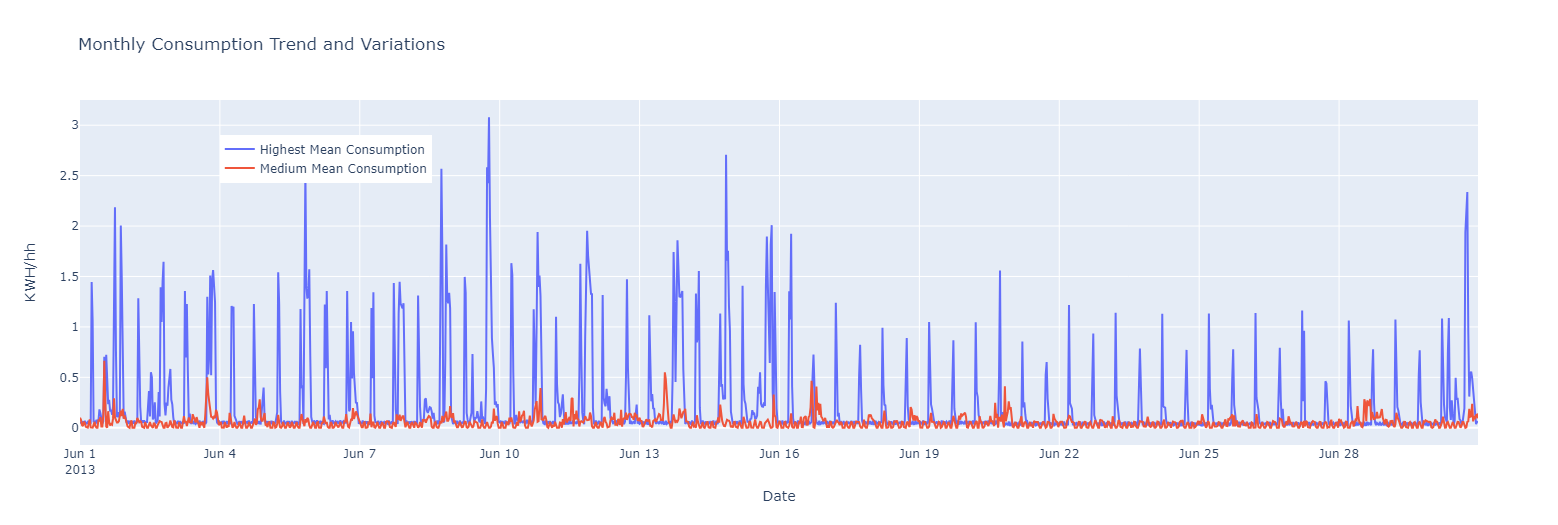}
        \caption{Comparison of daily consumption over a month: high consumption household vs average consumption household}
        \label{EP2}
    \label{fig:energy_profiles}
    \vspace{-5mm}
\end{figure}

First, we examined the heterogeneity of the consumption data and analyzed it to understand fluctuations in the energy consumption patterns of consumers across the year and to determine any seasonal variations or long-term trends.
Figure~\ref{subfig:p3} illustrates the monthly total consumption trend for a random household across the year, indicating that January, February, and December tend to have higher consumption compared to other months. 
Similarly, figure~\ref{EP2} shows the daily consumption of one high-consumption consumer and one medium-consumption consumer, capturing the significant variations and heterogeneity present in the dataset.  
We use the average consumption for each household across the entire dataset to identify outliers with very high or very low average half-hourly consumption. Based on our observations, households with an average energy consumption per half-hourly (hh) interval less than 0.09 kWh/hh (nearly unused) or greater than 1.35 kWh/hh are marked as outliers and filtered out of the dataset. 
Eliminating these outliers reduces the number of households from 5547 to 4672 and removes any bias that may be incurred when training the model with the outlier data points.


\subsection{Clustering of smart meter-based energy consumption data}

\begin{figure}[t!]
\centering
\includegraphics[width=0.42\textwidth]{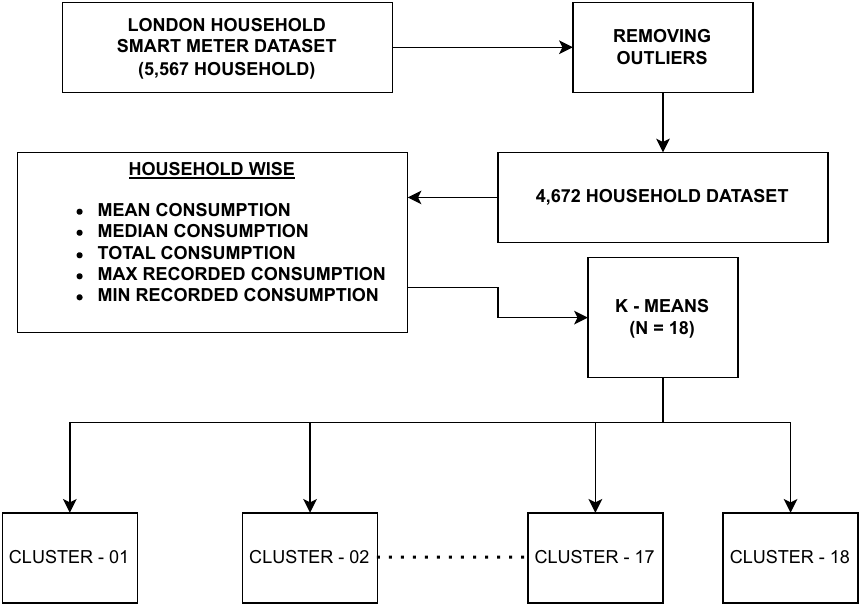}
\caption{Top to Cluster level Simulation Architecture}
\label{FD-1}
\end{figure}

Clustering the training data has shown to provide better generalisability for deep learning models when used for load prediction~\cite{bandara2020forecasting,quilumba2014using}.
With federated models, clustering is particularly effective as it allows clients to be organized into smaller clusters based on similar consumption and their locality. 
This also improves the aggregation phase in the substation where local models from similarly performing households have a better probability of achieving an unbiased aggregation to arrive at the new global model.
For our approach, we use statistical information about energy consumption to organize households into 18 groups using K-Means clustering. 
We use mean energy consumption, median consumption, total energy consumption, maximum recorded consumption, and minimum recorded consumption parameters to guide the clustering process, similar to the approach in~\cite{savi2021short}.
Clustering also takes into account the area locality for the grouping to ensure that they can be grouped under a global server that serves this area (e.g., a substation). So, in our setup, each cluster would indicate an area within London, comprising households of similar electric consumption levels with a substation (global server), feeding to all these households.   
Figure~\ref{FD-1} illustrates our constructed framework up to the cluster level.

\subsection{Federated Training and Testing Data sets Creation}

The entire dataset, indexed by household tags, is subsequently partitioned into training and test data. 
For every household, data from Nov 2011 to Feb 2013 was used for training and Validation purposes, while, the data from Feb 2013 to Feb 2014, was used for testing purposes, thereby achieving a 60:40 split between training and test data. 

\subsubsection{Input Window and Output Prediction}

To account for the time-series nature of load forecasting, we use a sliding window scheme to capture the short history of recent consumption data, which forms the input to the deep learning model. 
For our experiments, we evaluated different window sizes using training error at a fixed number of epochs to arrive at a size of 336, which captures an entire week of consumption at a half-hour resolution. 
We infer that by considering a longer time span that includes weekdays and weekends, we can capture and analyze the trends and patterns that emerge throughout a complete week.

\subsubsection{Model architecture and TensorFlow Federated Wrappers}

For building our training and inference flows (simulation), we utilise the Python-based TensorFlow-Federated (TFF) library and its associated frameworks. 
TFF wrapper converts the windowed input/output data into a Python dictionary for use with the TFF framework. 
For our evaluation, we simulate the learning of each individual household and its aggregation at the cluster using the TFF libraries. 
Unique tags associated with each household are used to access individual data streams and clusters during the training and evaluation phase. 

For our model, we chose to evaluate feed-forward neural networks for their simplicity and computational efficiency for each of the clients. 
After evaluating multiple architectures, we settled on a 4-layer model with 2 hidden layers. 
The model follows [16,8,4,1] shape with rectified linear unit (ReLU) activation functions at each layer. 
We used RMSE and MAE as the loss functions and metrics for our training and all three (RMSE, MAE and MAPE) for evaluation. 
The learning rate was set to 0.01 at each client site.
The final model architecture has 5569 trainable parameters, making it ideal for training and performing inference on low-cost microcontroller-based computing nodes (like smart meters) while having sufficient parameters to learn crucial patterns and trends in the data stream. 

\section{CASE STUDY}

\subsection{Case Settings}

We use the following hyper-parameters for training the model: batch size $=$ 12, global server learning rate $=$ 1 and number of federated rounds $=$ 20. 
We present a selection of results from individual clusters representing the spectrum of consumption levels that are available in the dataset in addition to the aggregate performance across all clusters.
For every cluster, two types of households are selected for evaluating our model performance: 
\begin{enumerate}
    \item Households that received all global model updates.
    \item Households that did not receive some updates or all updates during training.
\end{enumerate}
This differentiation factors in non-i.i.d conditions that may be present in a real-world condition. 
Additionally, within the group of households that did not receive some updates, we further divided them into inter-cluster examples, including high-consumption and low-consumption examples for observing how the forecasting accuracy's varies across different households. 

\subsection{Results}

\begin{table}[b!]
  \centering
  \caption{Monthly load forecasting RMSE: comparison between different cluster 08 households}
  \label{tab:T2}
  \scalebox{0.9}{
  \begin{tabular}{cccc}
    \hline
    \multirow{2}{*}{Month} & \multicolumn{3}{c}{Client Type (RMSE)} \\
    \cline{2-4}
    & Mod(All) & High(Non-i.i.d) & Low(Non-i.i.d) \\
    \hline
    January & 0.0636 & 0.2507 & 0.0778 \\
    February & 0.0633 & 0.2581 & 0.0665 \\
    March & 0.0623 & 0.2875 & 0.0760 \\
    April & 0.0544 & 0.1531 & 0.0797 \\
    May & 0.0570 & 0.0866 & 0.0740 \\
    June & 0.0621 & 0.0583 & 0.0752 \\
    July & 0.0638 & 0.1690 & 0.7425 \\
    August & 0.0662 & 0.2217 & 0.7119 \\
    September & 0.0528 & 0.1417 & 0.0730 \\
    October & 0.0611 & 0.1896 & 0.0695 \\
    November & 0.0539 & 0.2505 & 0.0684 \\
    December & 0.0650 & 0.2484 & 0.0787 \\
    \hline
  \end{tabular}}

\end{table}

Table \ref{tab:T2} presents the RMSE values averaged across different types of households within a specific cluster from our experiments. 
It can be observed that the model offers better (average) prediction performance for households that received all the updates (i.i.d condition) compared to those that missed out on some of the updates. 
Additionally, this demonstrates that in the worse scenario when a house skip updates, the approach is able to provide reasonably accurate predictions.

\begin{figure*}[t!]
    \centering
    \sbox{\bigpicturebox}{%
    \subfloat[Cluster 08\label{fig:stdf1}]{%
        \includegraphics[width=0.55\linewidth]{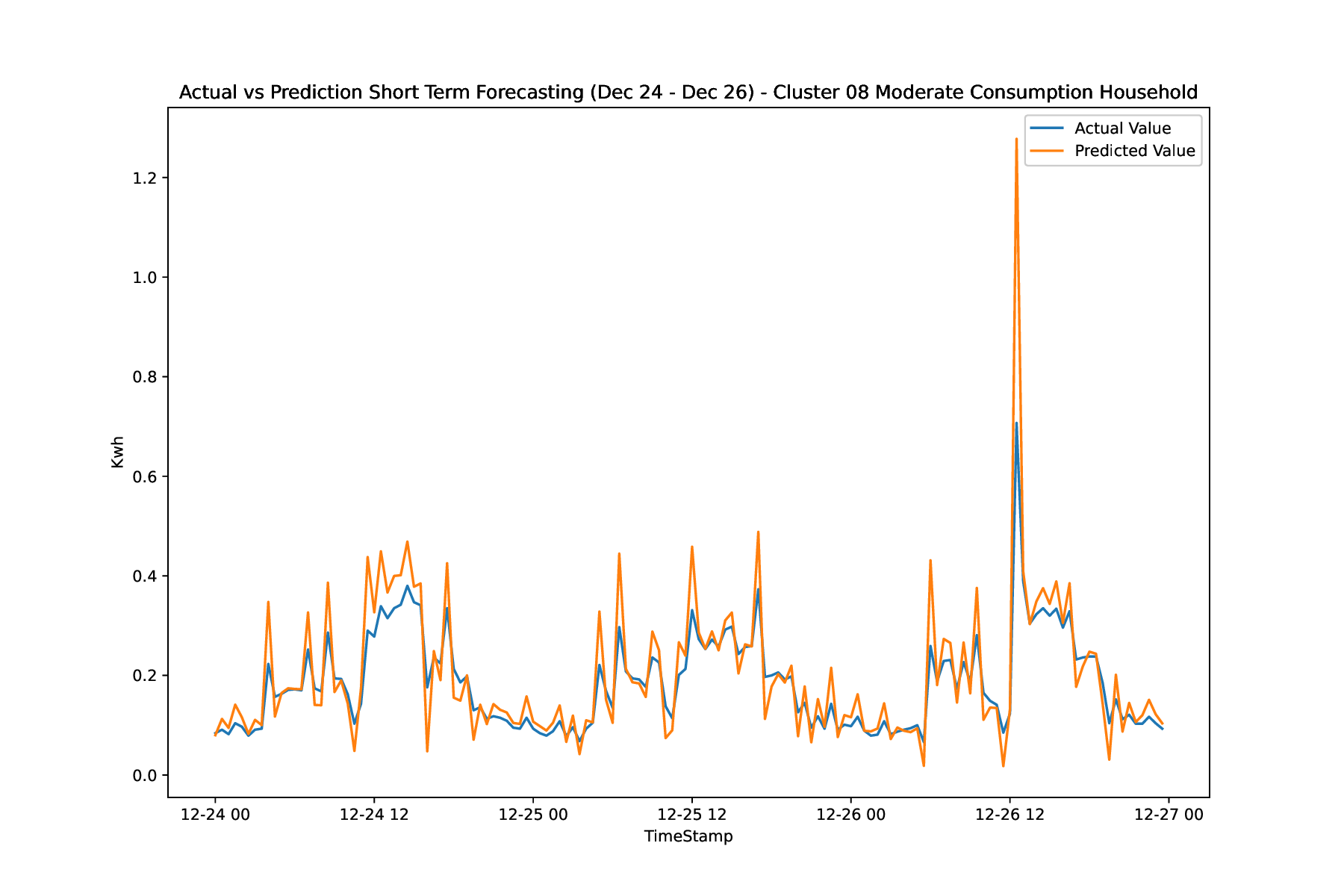}%
        }
    }    
    \usebox{\bigpicturebox}\hfill
    \begin{minipage}[b][\ht\bigpicturebox][b]{.45\textwidth}
    \subfloat[Cluster 04\label{fig:stdf2}]{%
        \includegraphics[width=0.45\linewidth]{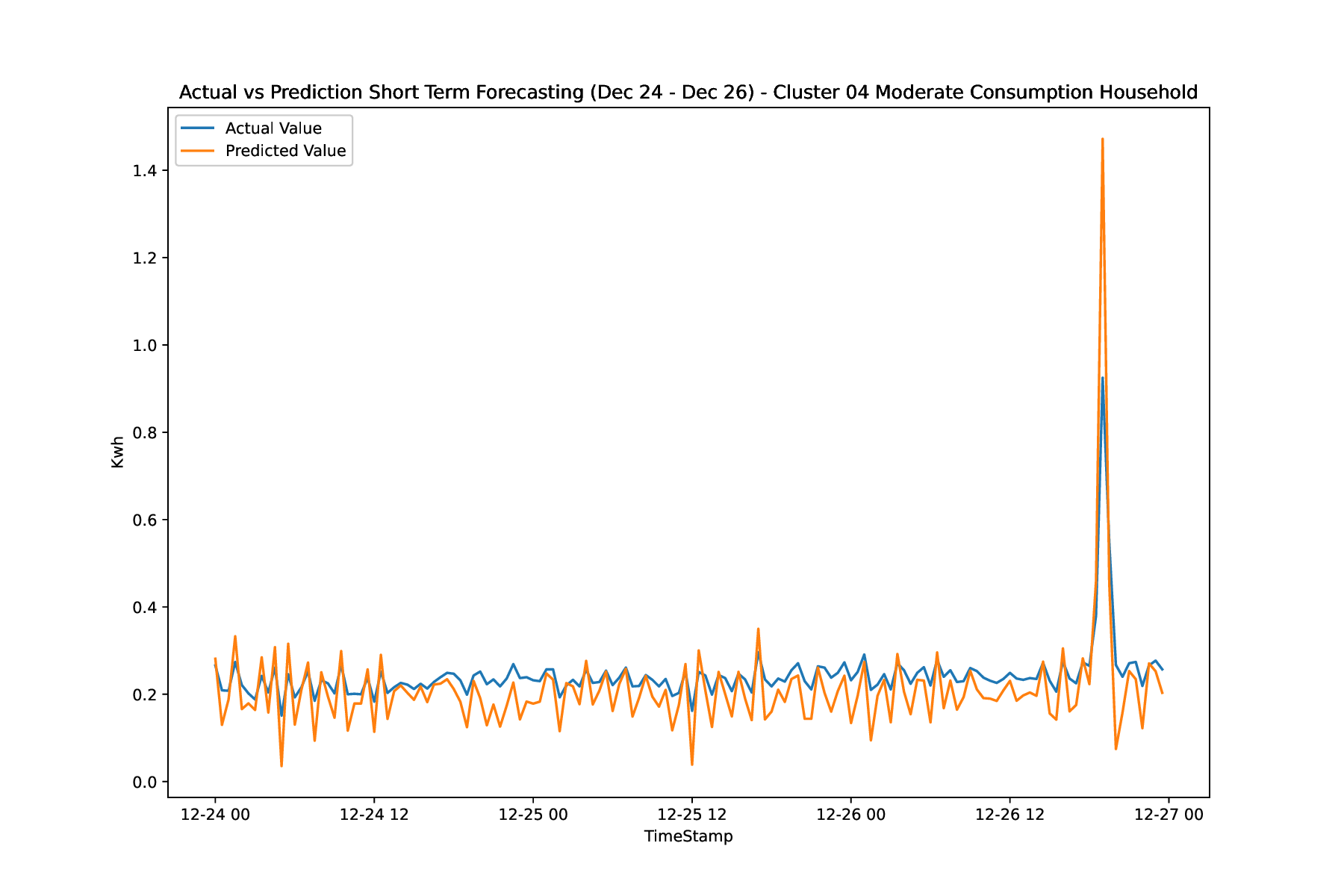}
    }
    \hfill
    \subfloat[Cluster 18\label{fig:stdf3}]{%
        \includegraphics[width=0.45\linewidth]{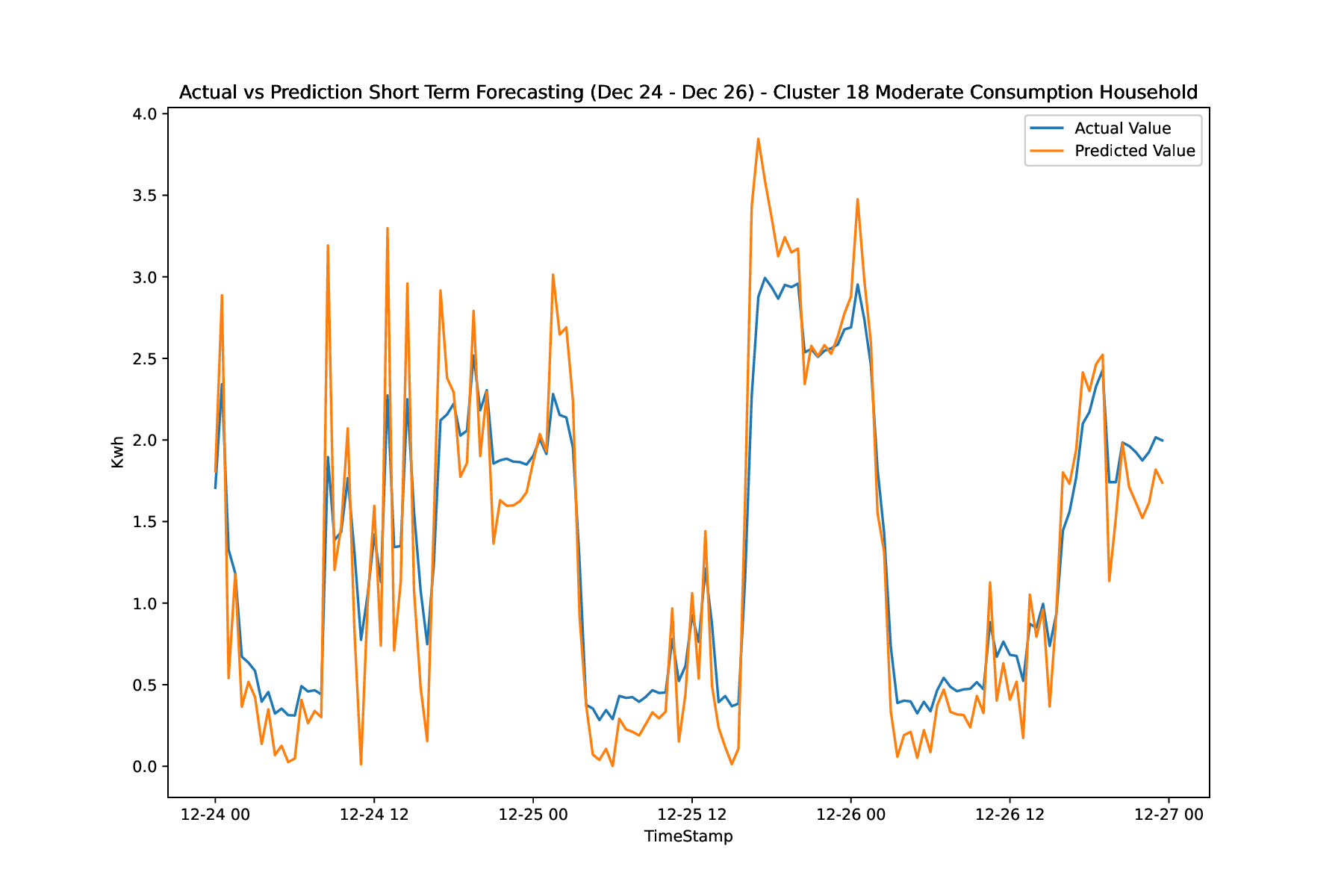}
    }
    \vfill
    \subfloat[Cluster 07\label{fig:stdf5}]{%
    %
        \includegraphics[width=0.45\linewidth]{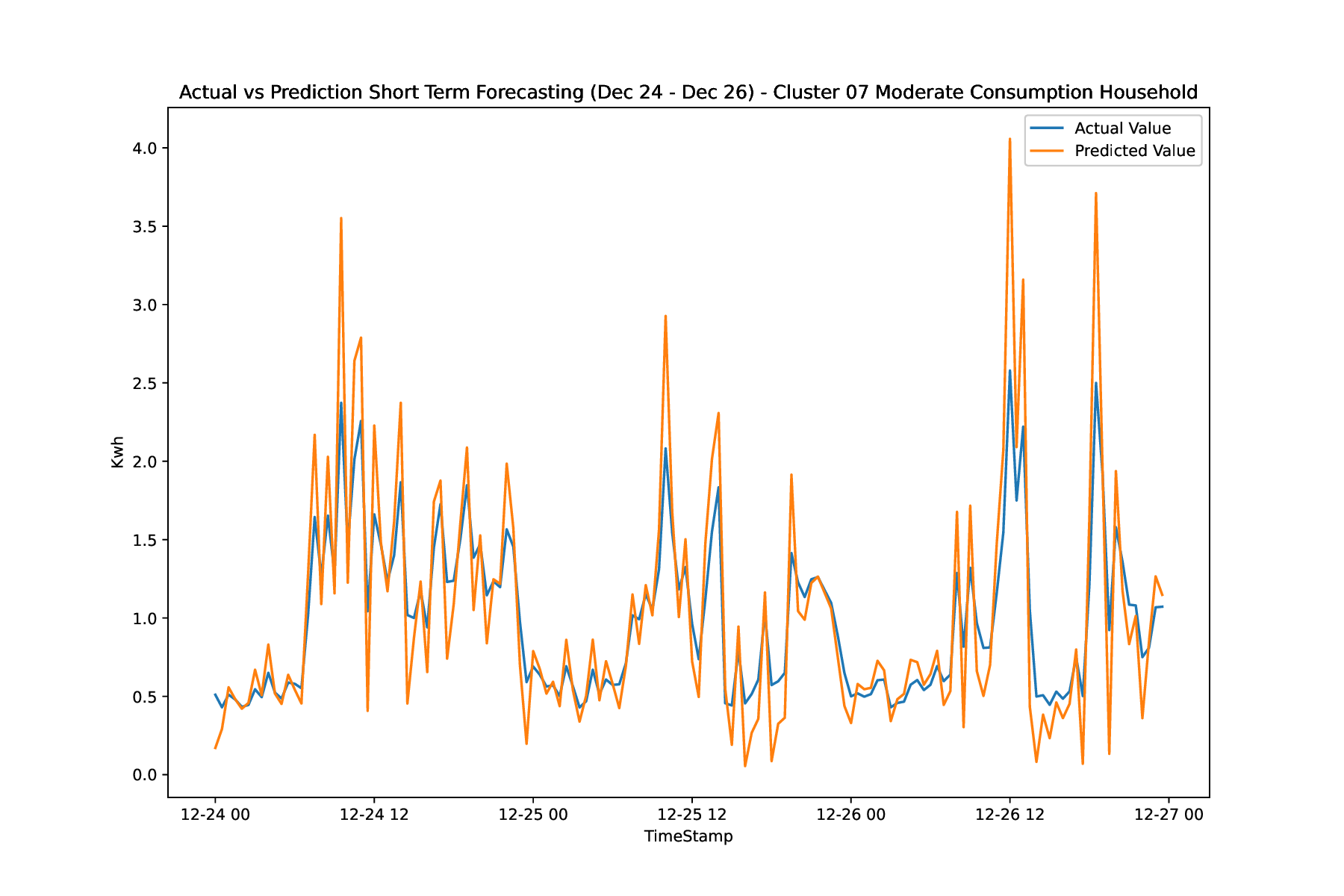}
    }
    \hfill
    \subfloat[Cluster 09\label{fig:stdf5}]{%
         \includegraphics[width=0.45\linewidth]{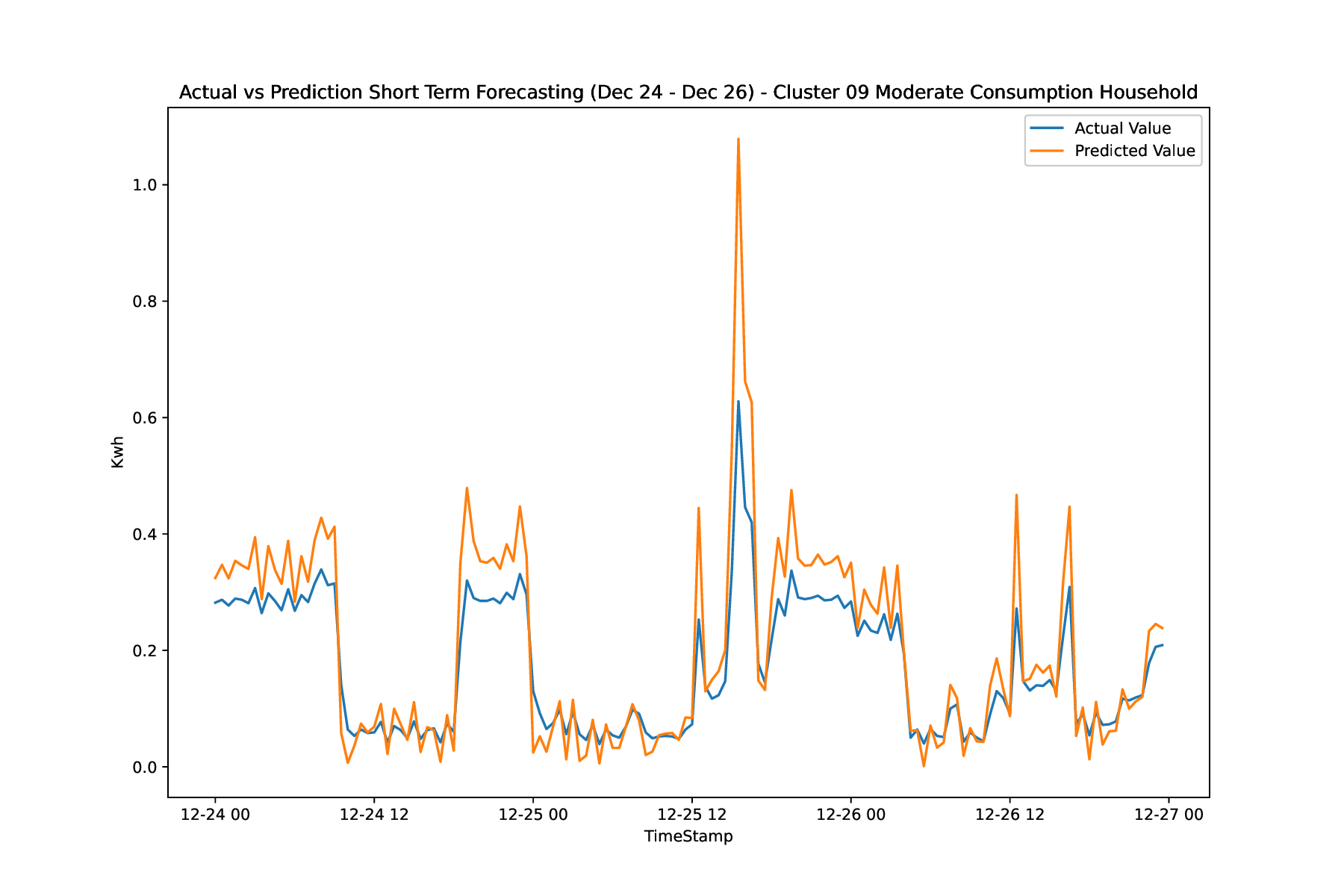}
    }
    \end{minipage}
    \caption{Short-term daily forecasting (24\textsuperscript{th} -- 26\textsuperscript{th} of December) across a selection of clusters}
    \label{fig:stdf}
    \vspace{-5mm}
\end{figure*}

We also examined the performance across festive periods where we observe significant differences in daily consumption while analyzing the dataset. 
The prediction results around a three-day window around Christmas from the 24\textsuperscript{th} of December to the 26\textsuperscript{th} of December is plotted in figure~\ref{fig:stdf} for different clusters. 
The general consumption data is well-aligned with the model's prediction, as can be seen from the figure across all clusters. 
While the sudden peaks are not accurately predicted by the model, the predicted consumption is on the lower side than the actual consumption, making it a reliable baseline for distribution system operators and grid operators. 

\subsubsection{Performance compared with Centralized Setup}

\begin{figure}[htbp]
    \centering
    \includegraphics[width=0.43\textwidth]{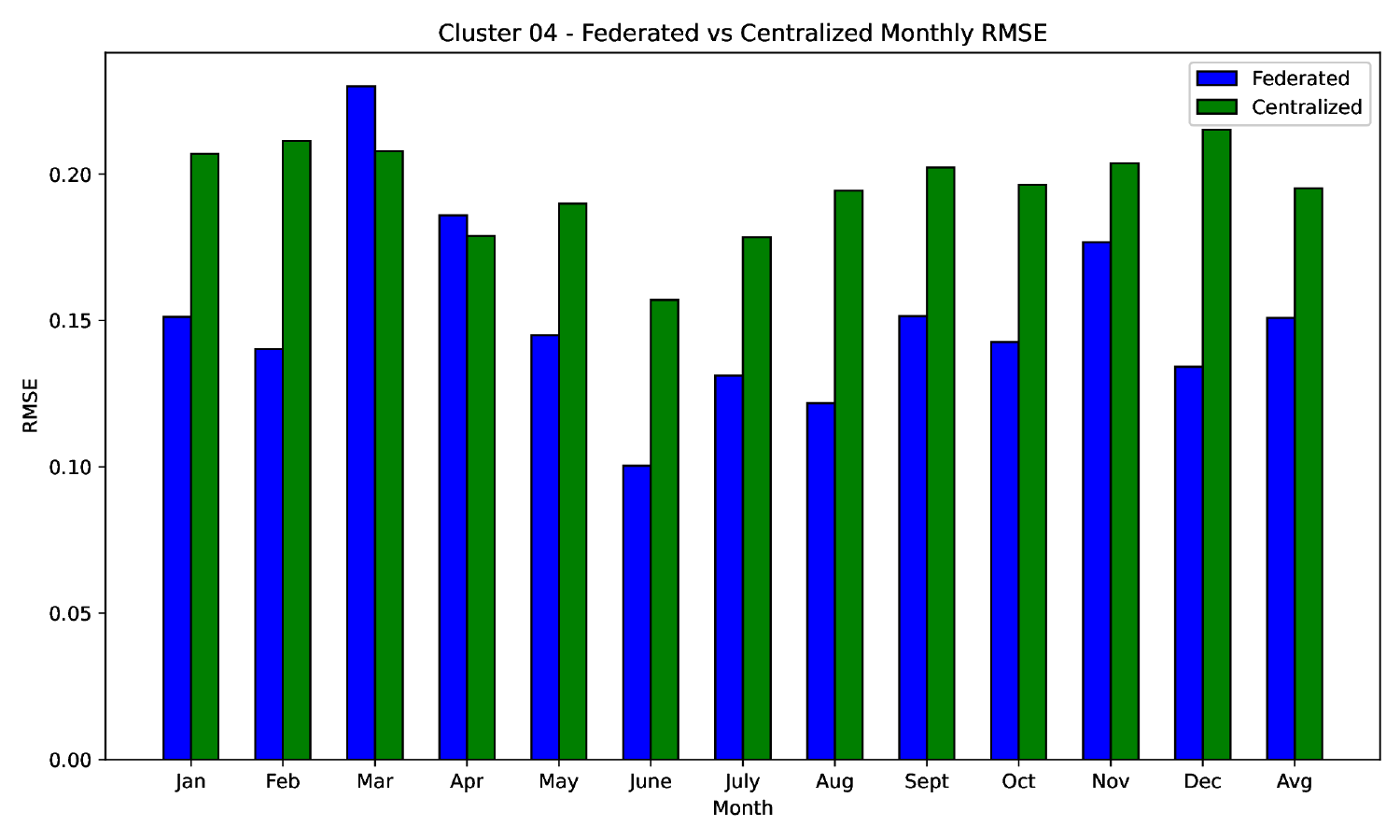}
    \caption{RMSE of centralized vs federated prediction for cluster 04}
    \label{fig:comp08}
\end{figure}

We evaluate the use of our lightweight model in a purely centralized setup to see the effectiveness of distributed learning in comparison to a global learning approach using a model of similar complexity. 
For the centralized setup, we assume that all consumption data is made available to the central server and that all clients participate at a consistent rate (set to 3$\times$ higher) compared to our federated learning scheme. 
The clustering approach is maintained in the centralized setup to ensure fairness in both cases. 
Both setups use 20 rounds of training - FL using training locally at the clients, while the centralized model trains for 20 epochs at the central server. 
Figure~\ref{fig:comp08} shows the average prediction error (RMSE) for the entire year for a single cluster (cluster 04) across different months. 
The results show that the federated model achieves better performance at the same model complexity.

\subsubsection{Performance comparison against other research works}

We compare our load forecasting model with the federated architecture used in~\cite{savi2021short}, using average monthly RMSE across various clusters. 
We include households with high, low and moderate electric energy consumption.
For our simulation, the number of participating clients was set to be less than 50\% of the clients used in the competing method~\cite{savi2021short} for each cluster. 
Although an increase in the number of participating nodes does not guarantee improvement in prediction performance, it should potentially lead to a better global model based on the consumption profile of the additionally included nodes.
Additionally, the work in~\cite{savi2021short} uses an LSTM-based model with 7505 parameters compared to our 4-layer feed-forward model with only 5569 learnable parameters (and thus lower communication costs during the model aggregation phase).

\begin{table}[t!]

\centering
\caption{Comparison of average RMSE of all clusters between our 4-layer feed-forward model and the LSTM-based FL model in~\cite{savi2021short}}
\label{Tab:comp}
\scalebox{0.9}{
\begin{tabular}{lcc}
\hline
\textbf{Month} & \textbf{Our Proposed Model (RMSE)} & \textbf{M. Savi Model (RMSE)} \\
\hline
January & 0.2053 & 0.1463 \\
February & 0.1626 & 0.153 \\
March & 0.2176 & 0.1535 \\
April & 0.1800 & 0.1345 \\
May & 0.1657 & 0.1304 \\
June & 0.1434 & 0.1259 \\
July & 0.1685 & 0.1057 \\
August & 0.1526 & 0.1106 \\
September & 0.1581 & 0.1302 \\
October & 0.1554 & 0.1335 \\
November & 0.1729 & 0.1471 \\
December & 0.1728 & 0.1328 \\
\hline
\end{tabular}}
\end{table}

The table~\ref{Tab:comp}) shows the results of our simulation capturing the average monthly RMSE of the models across all clusters over the entire year in comparison with the results reported in~\cite{savi2021short}.
Despite using 50\% fewer participating nodes in each cluster and 20\% fewer computational requirements for the local model, our model is able to achieve an average RMSE of 0.17 compared to 0.14 of the LSTM model. 
While RMSE does not fully capture the prediction performance in this case, we believe that the trade-off is in line with the lower computational complexity, communication overhead and participation ratio in the network.

We also compare our results against the work in~\cite{taik2020electrical}, where the model uses a participation ratio of 36\% compared to the 10\% in our method. 
Similar to the previous work, the work in~\cite{taik2020electrical} also uses an LSTM-based model with customized local model improvement at each round, with only 5 federated rounds compared to our 20. 
Our results, shown in table~\ref{Tab:comparison}, show that our approach achieves significantly better average prediction accuracy (in terms of MAPE) compared to the results reported in~\cite{taik2020electrical}. 
Additionally, we observe that reducing the number of federated rounds from 20 to 5 increases the average error across all clusters to increase by nearly 4\% in our case, which is still better than the results achieved by~\cite{taik2020electrical}.

\begin{table}[t!]
\centering
\caption{Comparison of Participation Ratio, Number of Rounds, and Average MAPE}
\label{Tab:comparison}
\scalebox{0.9}{
\begin{tabular}{lccc}
\toprule
Model & Household Ratio & Fed. Rounds & Avg MAPE \\
\midrule
Taik's Model & 36\% & 5 & 34.14\% \\
Our Proposed Model & 10.50\% & 20 & 22.01\% \\
\bottomrule
\end{tabular}}\vspace{-3mm}
\end{table}

\subsubsection{Energy consumption of the model on an IoT platform}

To quantify the energy overhead in performing model training and inference at each client (smart meter), we implement the training and inference on an Arduino Uno R4 WiFi platform, that uses an 8-bit processor. 
The model was coded using Ardiuno C and compiled to the device using the standard compiler flow. 
For evaluation, we simulate the condition that the model is training on the data for one day and predicted the half-hourly consumption for the following day, with the output transferred via serial port to a standard laptop. 
To measure the energy consumed, we monitor the power consumed by the Arduino device using a calibrated USB energy monitor and record the consumption of the model under idle, training and prediction phases. 
We observed that the model consumes 50\,mWh (milliwatt-hour) when averaged over the entire day on top of the idle power consumption of 440\,mWh, when performing training and inference, resulting in an overhead of 11\% on the Arduino platform. 
The result indicates that integration of our model will not incur significant energy consumption overhead in a similarly capable (or higher) smart meter hardware platform, making our approach ideally suited for this application.

\section{Conclusion}

In this work, we propose a federated learning-based approach for distributed load forecasting that uses a lightweight fully connected neural network at its core and test it on a real-world openly available energy consumption dataset. 
By combining this lightweight model with clustering and randomized client selection, we emulate real-world conditions in an electric distribution grid. 
Our approach learns locally at each client's smart meter with only the model updates passed up to the central entity, reducing the privacy risks associated with granular aggregation of energy consumption data. 
Our simulations show that the light-weight model when combined with clustering is able to generate accurate load forecasting, and provides comparable results against competing FL-based load prediction approaches, trading off inference accuracy ($\approx$21.4\% higher RMSE) with model complexity ($\approx$25.8\% lower parameters) in one case, while achieving $\approx$35.6\% better aggregate MAPE in another. 
We also show that training and inference using our model on an Arduino platform incurs an energy overhead of only 50\,mWh when averaged over multiple cycles, which is a fraction of the standalone power consumption for the platform, making it ideally suited for load forecasting in a decentralized setting.
In the future, we aim to show that decentralized learning can aid in improving the smart-grids stability and resilence without requiring aggregation of privacy-sensitive data.

\bibliographystyle{IEEEtran}
\bibliography{FL_DGupta}

\end{document}